\documentclass[letterpaper]{article} % DO NOT CHANGE THIS
\usepackage[preprint]{aaai2027}  % DO NOT CHANGE THIS
\usepackage[hyphens]{url}  % DO NOT CHANGE THIS
\usepackage{graphicx} % DO NOT CHANGE THIS
\usepackage{natbib}  % DO NOT CHANGE THIS AND DO NOT ADD ANY OPTIONS TO IT
\usepackage{caption} % DO NOT CHANGE THIS AND DO NOT ADD ANY OPTIONS TO IT
\usepackage{algorithm}
\usepackage{algorithmic}
\usepackage{amsmath}

\usepackage{newfloat}
\usepackage{listings}
\DeclareCaptionStyle{ruled}{labelfont=normalfont,labelsep=colon,strut=off} % DO NOT CHANGE THIS
\floatstyle{ruled}
\newfloat{listing}{tb}{lst}{}
\floatname{listing}{Listing}

\usepackage{booktabs}

\usepackage{array}

\newcolumntype{L}[1]{>{\raggedright\arraybackslash}p{#1}}
\newcolumntype{C}[1]{>{\centering\arraybackslash}p{#1}}
\title{\centering\raisebox{-1.2ex}{\includegraphics[height=2em]{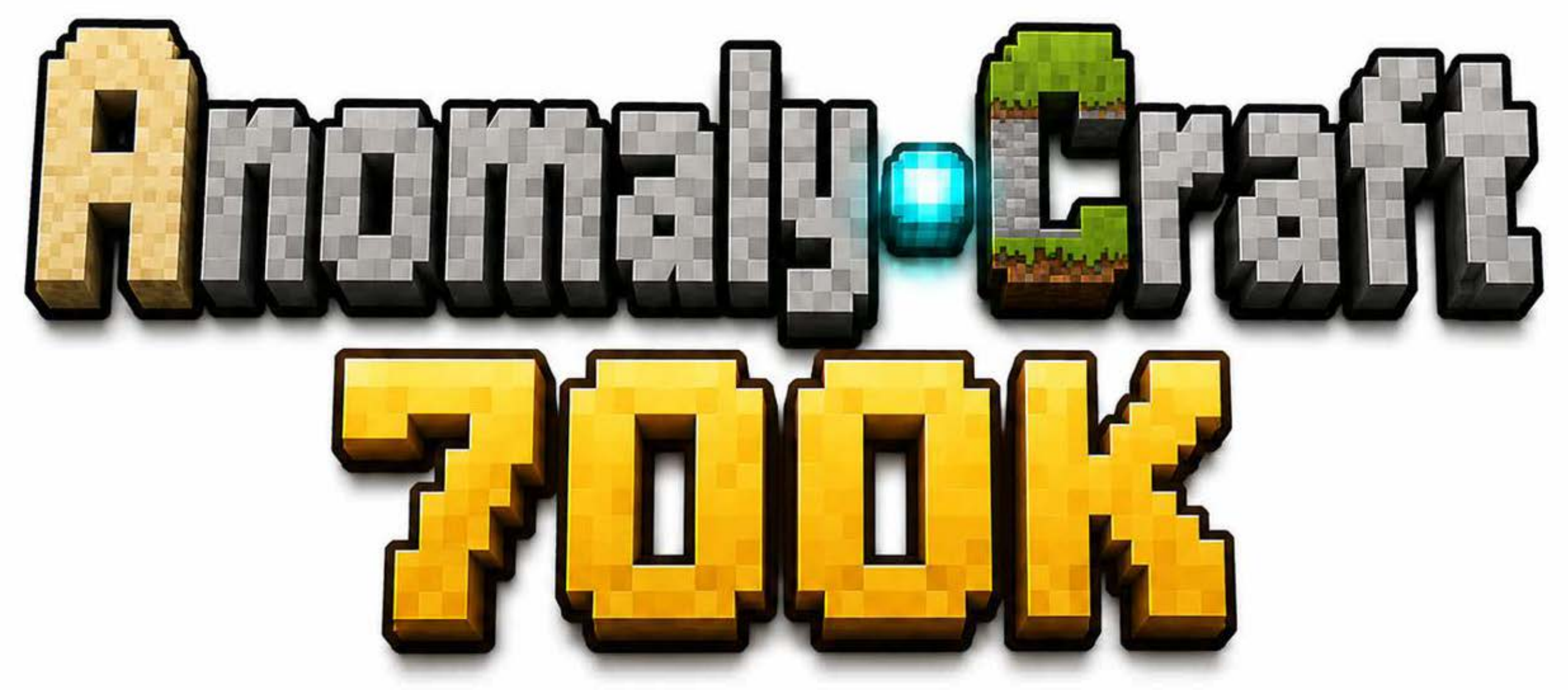}}AnomalyCraft-700K: Component-Level Controllable and Verifiable Synthetic Anomalies for Fine-Grained Video Anomaly Understanding}
\author{
    Yuzhou Long\textsuperscript{\rm 1},
    Haodong Zhang\textsuperscript{\rm 1},
    Yunpeng Yang\textsuperscript{\rm 1},
    Peng Wu\textsuperscript{\rm 1}\corresponding,
    Guansong Pang\textsuperscript{\rm 2}
}
\affiliations{
    \textsuperscript{\rm 1}Northwestern Polytechnical University 
    \quad
    \textsuperscript{\rm 2}Singapore Management University \\
    Project Page: \url{https://github.com/Eagen-l/AnomalyCraft}
    
}

\begin{document}

\maketitle

\begin{abstract}
Progress in video anomaly understanding (VAU) has long been limited by inherent deficiencies of real-world anomaly videos, which are hard to collect and offer little control over their content. Synthetic anomaly approaches partially alleviate data scarcity, 
yet their generation remains largely controlled at the category or prompt level. They also lack component-level verification of video-text consistency and provide insufficient hard normal samples near the normal-anomaly boundary. 
To address this, we present AnomalyCraft-700K, a component-controllable synthetic anomaly dataset for fine-grained VAU, containing over 40K videos and over 700K task-level textual annotations. From fine-grained semantic components and a progressive three-stage pipeline, we craft anomaly events that are richly detailed, semantically controlled, and temporally structured, and additionally construct per-category hard normal samples to prompt the model to discriminate based on anomaly semantics rather than surface visual cues. Moreover, using the components as verification units, AnomalyCraft-700K further performs component-wise correction of video–text discrepancies introduced during generation, providing reliable annotations with verified cross-modal alignment for six tasks that progress from anomaly detection, through anomaly retrieval and captioning, to fine-grained anomaly reasoning. Evaluations of widely used methods under both traditional and MLLM-based protocols demonstrate that AnomalyCraft-700K serves as an effective source of supervision, from anomaly detection to fine-grained anomaly understanding.
\end{abstract}

\section{Introduction}

The evolution of video anomaly understanding (VAU) is closely tied to the evolution of available datasets. Early benchmarks such as UCSD Ped~\cite{li2013anomaly} and CUHK Avenue~\cite{lu2013abnormal} mainly contain a few fixed scenes, limited anomaly types, and often staged abnormal behaviors. They establish the classical one-class video anomaly detection (VAD) setting, where models learn scene-specific normality and detect deviations at test time. Later benchmarks, including UCF-Crime~\cite{sultani2018real} and XD-Violence~\cite{wu2020not}, substantially expand data scale, scene diversity, and anomaly coverage through real-world surveillance and online videos. They shift VAD from normality modeling in constrained environments toward anomaly recognition in complex, untrimmed videos. 
Synthetic data, such as UBnormal~\cite{acsintoae2022ubnormal}, further reshapes this progression by allowing abnormal events to be deliberately constructed rather than passively collected. However, its videos remain based on simplified scenes and scripted actions, leaving a substantial gap in visual realism and contextual complexity relative to real-world anomalies. 
Across these benchmarks, supervision is still largely detection-oriented, compressing events with distinct participants, interactions, and temporal processes into binary or category-level labels.

\begin{figure}[t]
\centering
\includegraphics[
    width=\columnwidth
]{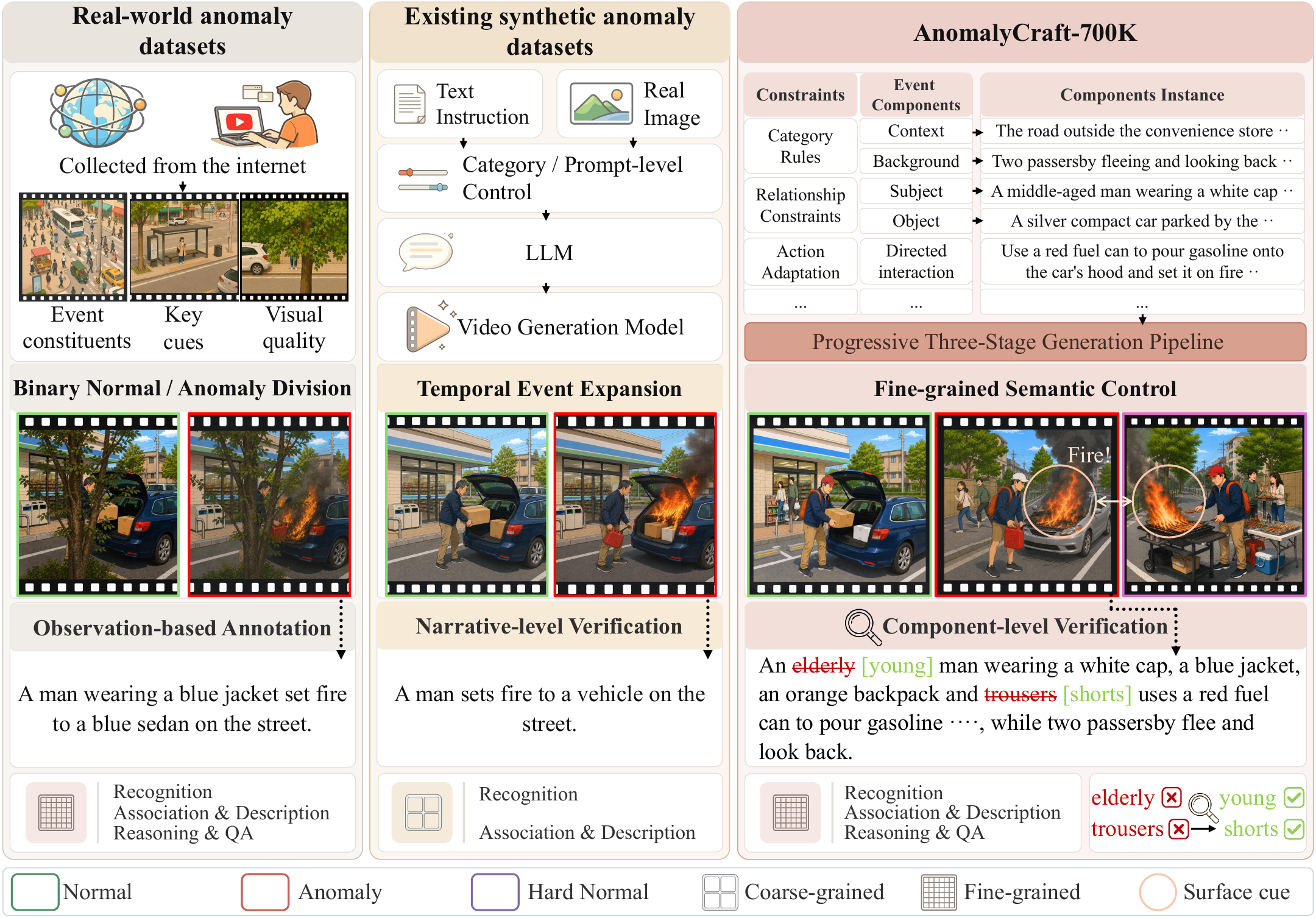}
\caption{Comparison of AnomalyCraft-700K with existing anomaly datasets.}
\label{Fig1_draft}
\end{figure}

The emergence of vision-language models (VLMs) further extends the field toward video anomaly understanding. Recent benchmarks, such as UCA~\cite{yuan2024towards}, Holmes-VAU~\cite{zhang2025holmes}, and VAU-R1~\cite{zhu2025vaur1advancingvideoanomaly}, augment existing anomaly datasets with language supervision to support fine-grained anomaly understanding. These efforts enable models to address not only whether an anomaly occurs, but also what happens and why it is abnormal. However, such annotations are added retrospectively to videos whose content is already fixed. Relevant participants, objects, interactions, and category-defining cues are therefore neither planned nor systematically covered. Manual annotation is costly, while generic automatic annotation tends to focus on salient evidence and overlook subtle but distinctive details.

Recent video generation models provide a new route toward scalable anomaly data construction. SVTA~\cite{yang2025scalablevideoanomalyretrieval} demonstrates the value of synthetic video-text pairs for cross-modal anomaly retrieval, while Pistachio~\cite{li2026pistachiosyntheticbalancedlongform} further improves scene diversity, temporal continuity, and multi-stage anomaly narratives through storyline-conditioned generation. Yet scalable synthesis does not by itself ensure fine-grained semantic controllability or reliable supervision. Existing control is mainly exercised over scenes, anomaly categories, and storyline segments, while the semantic constituents that determine how an anomaly is realized remain insufficiently parameterized. 
Moreover, generation prompts are often inherited as annotations, either directly or after event-level verification, leaving component-level semantic discrepancies largely unchecked.
Normal samples are also generally introduced to enrich behavioral diversity rather than to form category-specific semantic boundaries with visually similar anomalies.

In this paper, we introduce AnomalyCraft-700K, a component-controllable synthetic dataset for fine-grained video anomaly understanding. It advances anomaly synthesis from category- and storyline-level generation to fine-grained semantic construction, while further aligning generation intent with the content actually realized in the video. 
Specifically, AnomalyCraft-700K decomposes each anomaly event into five core components, including context, background, subject, object, and directed interaction, together with category-specific attributes. A progressive three-stage pipeline gradually relaxes non-essential constraints while preserving the defining anomaly semantics, balancing precise controllability with event diversity and temporal structure. The same semantic components are further reused as verification units, allowing component-to-video discrepancies to be identified and manually corrected rather than directly treating generation prompts as annotations. We also construct category-specific hard normal counterparts that preserve anomaly-like visual cues while remaining semantically normal, encouraging models to distinguish anomalies through event semantics rather than superficial appearance. 
To support comprehensive anomaly understanding, AnomalyCraft-700K provides six tasks across three capability dimensions: anomaly detection, anomaly retrieval and captioning, and fine-grained anomaly reasoning. To our knowledge, it is the first synthetic anomaly dataset to jointly support controllable construction at the component level and explicit component-aware video-text verification. It contains 40,100 videos, more than 3 million frames, and over 700K task-level multimodal supervision instances, enabling evaluation from conventional anomaly detection to fine-grained semantic reasoning.

\begin{itemize}
    \item We propose the first synthetic anomaly dataset with controllable fine-grained anomaly semantics, achieved by composing structured semantic components through a three-stage pipeline that mitigates the limited anomaly priors of general-purpose video generators.
    \item We turn the original generation-time components into targeted units, checking and correcting component-level discrepancies between the intended semantic components and their visual realization item by item, pushing cross-modal alignment verification from the narrative level down to the component level and providing downstream tasks with verified fine-grained alignment annotations.
    \item Guided by the surface visual cues of each anomaly category, we construct boundary samples that are visually close to anomalies yet semantically normal, so as to prompt the model to discriminate based on anomaly semantics rather than surface visual cues.
    \item We provide unified supervision for six tasks spanning three capability dimensions, supporting a complete evaluation from anomaly detection to fine-grained semantic understanding.
\end{itemize}

\section{Related Work}
\subsection{Video Anomaly Detection}

Traditional VAD primarily formulates anomaly detection as visual anomaly discrimination. Early approaches ~\cite{wu2019deep} typically adopt one-class learning to model normal patterns from training videos, whereas recent weakly supervised methods~\cite{li2022self, park2023normality,lv2023unbiased,chen2026mgfn++,chen2024prompt,xu2025discriminative,majhi2025just,acharya2026road,wang2026weaklysupervisedvideoanomaly} learn frame-level binary discrimination from coarse video-level labels. With the emergence of vision-language pretraining, methods~\cite{yang2024text,wu2024open,wu2024weakly,wang2025federated,ye2025vera,li2025anomize,yin2026learning,zou2026unlocking,zhu2026alert} such as VadCLIP~\cite{wu2024vadclip} further adapt CLIP~\cite{radford2021learning} to VAD, exploiting vision-language alignment to improve anomaly detection while still operating within a classification-oriented paradigm. Despite these advances, conventional VAD remains centered on detection, with limited ability to capture the fine-grained semantics of anomalous events. 
Recent VLMs shift VAD from score prediction toward semantic reasoning. Rather than directly outputting anomaly scores, these methods first interpret what happens and why it may be abnormal, and then infer whether the event is anomalous. Holmes-VAU introduces language-based anomaly judgment into video anomaly understanding, while training-free approaches such as LAVAD~\cite{zanella2024harnessing} and VADTree~\cite{li2026vadtree} further explore reasoning-based VAD without task-specific model training.

\subsection{Video Anomaly Retrieval and Captioning}

Video anomaly retrieval and captioning aim to establish fine-grained semantic correspondence between anomalous video content and natural-language descriptions~\cite{luo2022clip4clip,ma2022x,liu2022ts2,wang2023unified,jeong2025learning,lan2025hybrid,bang2026beyond,chen2026eaglenetenergyawarefinegrainedrelationship}. Retrieval methods~\cite{long2025single} localize relevant anomalous events in videos according to textual queries, while captioning methods generate language descriptions of observed events. UCA extends UCF-Crime with temporally grounded event descriptions, supporting both language-based localization and caption generation for surveillance videos. ALAN~\cite{wu2024toward} and VarCMP~\cite{wu2025varcmp} adapt cross-modal pretrained models to capture fine-grained correspondence between long videos and textual queries. To overcome the limited scale of real-world video-text pairs, Yang et al.~\cite{yang2025scalablevideoanomalyretrieval} introduce SVTA, a large-scale synthetic benchmark in which textual descriptions guide anomaly video generation. This work demonstrates the potential of generation models for scaling anomaly-focused video-text alignment. However, SVTA mainly targets retrieval, with video-level descriptions serving as paired queries for generated videos, while richer forms of anomaly understanding, such as structured description, fine-grained reasoning, and question answering, remain beyond its primary scope.

\subsection{Fine-grained Video Anomaly  Understanding }

Recent video understanding research has expanded from caption generation toward fine-grained, reasoning-oriented supervision. For example, OmniVideo-100K~\cite{cai2026omnivideo100kdatasetaudiovisualreasoning} derives evidence-grounded question-answer pairs from audio-visual videos, and BusterX~\cite{wen2026busterxmllmpoweredaigeneratedvideo} formulates AI-generated video detection as a step-by-step visual reasoning task. In surveillance scenarios, Holmes-VAU constructs HIVAU-70K, a multi-granular dataset for anomaly description and analysis, while VAU-R1 organizes anomaly understanding into multiple-choice question answering, temporal grounding, and anomaly reasoning. These works enrich real-world videos with language supervision but remain constrained by the inherent long-tailed distribution and uncontrollable quality of real-world videos. Pistachio adopts a segment-wise generation-and-concatenation scheme to extend synthetic anomalies along the temporal axis into long-form, multi-stage sequences, and verifies text-video consistency at the narrative level. However, existing synthetic work still perform generation control and video–text verification primarily at the narrative level, without achieving controllable construction of fine-grained anomaly semantics at the component level or relieving the long-standing shortage of hard normal samples. In contrast, AnomalyCraft-700K performs both generation control and video–text verification at the component level. It also introduces category-specific hard normal samples and unified supervision for six anomaly understanding tasks.

\begin{figure*}[t]
\centering
\includegraphics[
    width=0.92\textwidth,
]{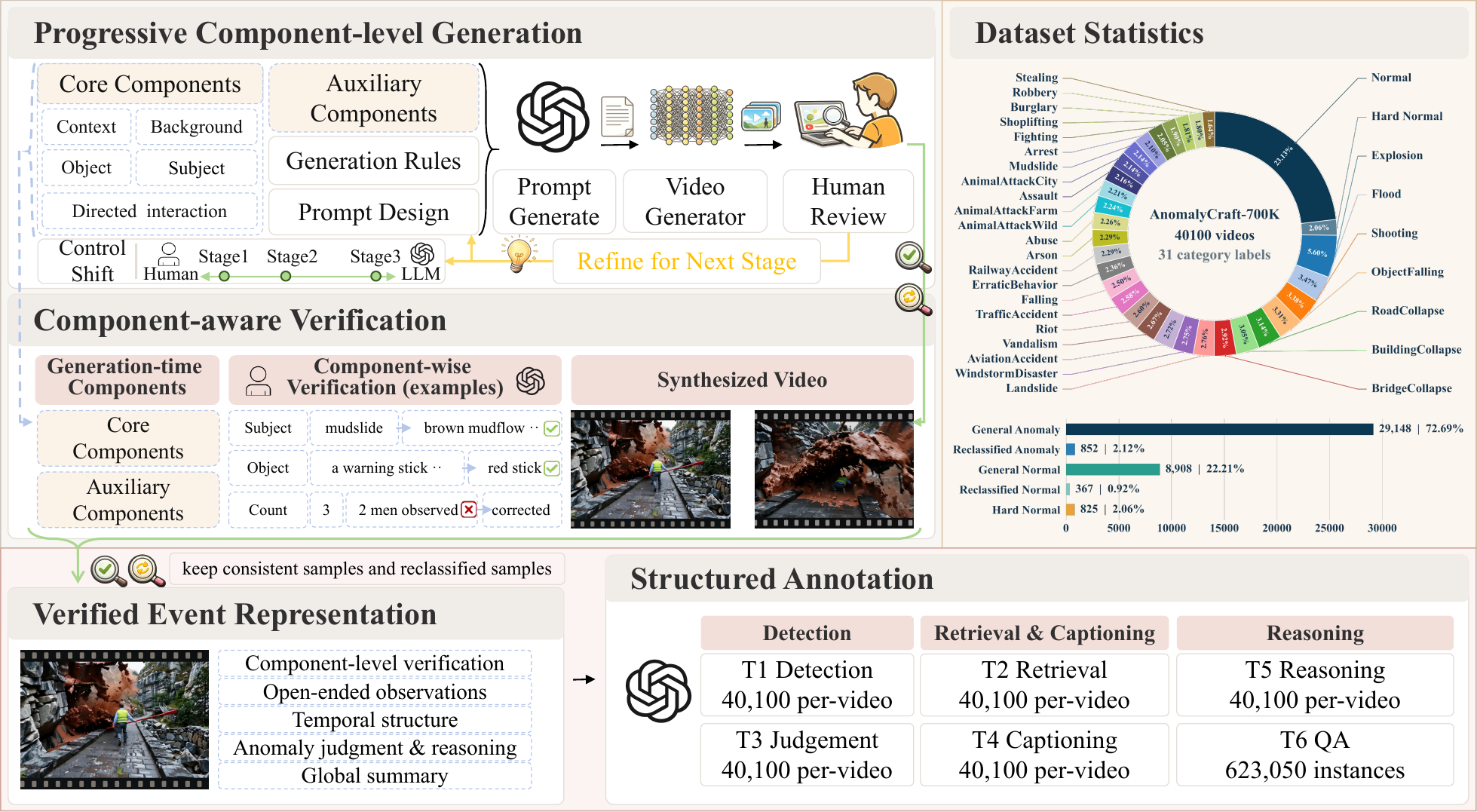}
\caption{Overview of our video anomaly dataset generation pipeline}
\label{fig:pipeline}
\end{figure*}

\section{AnomalyCraft-700K}

\subsection{Component-level Anomaly Representation}

Although recent video generation models achieve impressive visual realism, their compositional understanding of long-tailed anomalous events remains limited. Directly prompting a generative model with an anomaly category often produces incomplete or semantically ambiguous results. Instead of treating an anomaly as a holistic concept, we treat an anomalous event as a dynamic process involving many components, progressing over time and gradually deviating from the normal state, formalized as $V = (C, T)$, where $C$ denotes the semantic components of the event and $T$ describes their temporal evolution. 
Each event is decomposed into five core components:
\begin{equation}
\begin{aligned}
C = \bigl(
&\text{Context},
 \text{Background},
 \text{Subject},\\
&\qquad
 \text{Object},
 \text{Directed Interaction}
\bigr)
\end{aligned}
\end{equation}

These components jointly determine the anomaly semantics. Context and Background describe where the event occurs and the surrounding environment. Subject and Object specify the participating entities, while Directed Interaction characterizes the abnormal relation between them. Unlike existing synthetic datasets that mainly control anomaly categories or storylines, this representation explicitly parameterizes the semantic constituents that determine how an anomaly is instantiated.
Within this component framework, we further specify the value range of each component. By subject type, we construct 29 anomaly categories, of which 17 involve human or animal subjects (animate) and 12 have vehicle, natural-environment, or various object and facility subjects (inanimate). Following the characteristics of each category, we construct several subject–object pairings involving humans, animals, vehicles, objects or facilities, as well as cases without an explicit object, and specify the admissible directed interactions for particular pairings. To further constrain category semantics and expand diversity, we add category-specific components such as subject-object counts, relative positions, and interaction objects, together forming a comprehensive component space.

Since anomalies are dynamic processes rather than isolated moments, we further organize component evolution into four temporal stages:
\begin{equation}
T = \left(\text{Initial}, \text{Trigger}, \text{Escalation}, \text{Outcome}\right)
\end{equation} 
these stages describe how an event evolves from an initial state, through the onset and escalation of abnormal behavior, to its final outcome. This temporal representation provides explicit guidance for event progression and serves as the basis for the progressive generation pipeline introduced in the following section.

\subsection{Progressive Component-level Generation}

Although the proposed component representation explicitly specifies anomaly semantics, directly converting all components into a single prompt often leads to unstable generation. Current video generators cannot reliably instantiate all fine-grained semantic constraints simultaneously, causing missing components, incorrect interactions, or semantic drift. We therefore develop a progressive component-level generation pipeline that gradually relaxes generation constraints while preserving the defining anomaly semantics.

As illustrated in Figure~\ref{fig:pipeline}, the pipeline consists of three stages. Each stage follows the same generation loop: structured components are first converted into a generator-oriented description, a video is synthesized, and the generated results are reviewed to refine the construction strategy of the subsequent stage. Across stages, human control is progressively relaxed from complete specification of all semantic components to preserving only the defining anomaly semantics, while increasingly allowing the LLM to instantiate unspecified semantic details in the generation description. This progressive strategy balances controllability with event diversity and enables us to gradually explore the generation capability of the underlying video model.

Specifically, Stage I adopts fully specified component combinations, where every core and auxiliary component is explicitly determined before generation. This stage establishes reliable generation rules and identifies the generator's capability boundary for different anomaly components. Stage II gradually relaxes auxiliary semantic constraints while preserving the predefined anomaly semantics, allowing the LLM to enrich event details and increase scene diversity. Finally, Stage III retains only the core anomaly semantics, including the anomaly category, subject-object pairing, and directed interaction, leaving the remaining details to the LLM. This progressively enlarges the semantic composition space while maintaining category consistency and event plausibility.

The structured components are translated into generator-oriented descriptions following two simple principles. First, static components are organized before dynamic interactions, allowing the scene and participating entities to be established before the anomaly evolves. Second, a continuation signal describing the subsequent state of the subject and object is appended to stabilize the event semantics throughout the target video duration and reduce semantic drift near the end of generation. Together, these descriptions provide temporally coherent control signals that are better aligned with video generation than conventional human-oriented narratives.

By progressively relaxing non-essential constraints while preserving the defining anomaly semantics, the proposed pipeline effectively reconciles precise semantic control with generation diversity, laying the foundation for the subsequent component-aware verification process.

\begin{table*}[t]
\centering
{\small
\tabcolsep=2pt
\begin{tabular*}{\textwidth}{
@{\extracolsep{\fill}}
lcccccccl
@{}
}
\toprule
\raisebox{0.6em}{\textbf{Dataset}} &
\shortstack{\textbf{Anno.}\\\textbf{verification}} &
\shortstack{\textbf{Hard}\\\textbf{normal}} &
\raisebox{0.4em}{\textbf{\#Tasks}} &
\raisebox{0.6em}{\textbf{\#Anno.}} &
\raisebox{0.6em}{\textbf{\#Videos}} &
\raisebox{0.6em}{\textbf{\#Frames}} &
\raisebox{0.6em}{\textbf{\#Categories}} &
\raisebox{0.6em}{\textbf{Source}} \\
\midrule

UCA~\shortcite{yuan2024towards} &
-- &
No &
4 &
23,542 &
1,854 &
N/R &
13+1 &
Real \\

HIVAU-70K~\shortcite{zhang2025holmes} &
-- &
No &
2 &
72,325 &
5,443 &
N/R &
14+1 &
Real \\

VAU-Bench~\shortcite{zhu2025vaur1advancingvideoanomaly} &
-- &
No &
4 &
N/R &
4,602 &
N/R &
19+1 &
Real \\

SVTA~\shortcite{yang2025scalablevideoanomalyretrieval} &
-- &
No &
1 &
41,315 &
41,315 &
1.36M &
68+30 &
Synthesis \\

Pistachio~\shortcite{li2026pistachiosyntheticbalancedlongform} &
Narrative &
No &
2 &
N/R &
6,347 &
2.19M &
31+1 &
Synthesis \\

\textbf{AnomalyCraft-700K} &
\textbf{Component} &
\textbf{Yes} &
\textbf{6} &
\textbf{743,350} &
\textbf{40,100} &
\textbf{3.01M} &
\textbf{29+1} &
\textbf{Synthesis} \\

\bottomrule
\end{tabular*}
}
\caption{Comparison of video anomaly understanding datasets.}
\label{tab:dataset_comparison}
\end{table*}

\subsection{Component-aware Post-generation Verification}

Existing synthetic anomaly datasets generally inherit generation prompts as annotations, either directly or after narrative-level verification. However, component-level generation introduces much finer semantic constraints, making it insufficient to assume that every intended component is faithfully realized in the synthesized video. We therefore establish a component-aware verification process to ensure that all downstream annotations are derived from the generated video rather than the generation prompt. 
The verification process consists of four steps. We first perform event-level screening to remove low-quality samples and reclassify videos that depict a different anomaly category. Each predefined semantic component is then treated as an independent verification unit, and an MLLM determines whether it is correctly instantiated while simultaneously describing the content actually observed in the video. Open-ended observations are further used to capture visual information beyond the predefined components. Based on these results, semantic discrepancies are corrected by updating component values, revising anomaly categories, or discarding unclassifiable samples. Finally, human reviewers examine ambiguous cases and make final annotation decisions. 
The same verification protocol is applied throughout all three construction stages. As the generation pipeline progressively matures, the video retention rate increases from 81.56\% to 90.05\% and finally 95.55\%, providing evidence that the progressive pipeline yields increasingly reliable component realization across stages. 
Consequently, all downstream annotations are derived from verified video semantics rather than inherited generation prompts.

\subsection{Category-specific Hard Normal Construction}

In existing anomaly datasets, normal and anomalous samples often differ markedly in visual appearance, allowing models to rely on superficial cues. To reduce this shortcut, we introduce category-specific hard normal counterparts that preserve anomaly-like visual cues while remaining semantically normal. 
Specifically, we reuse the characteristic visual cues of each anomaly category and modify only the semantic factors that determine abnormality. Hard normal samples are constructed by altering one or more of five semantic dimensions, including authorization, occupational purpose, rule constraint, controlled process, and harmless outcome, while preserving the overall visual appearance of the event. For example, placing merchandise into a bag under staff supervision remains visually similar to Shoplifting but differs only in authorization. Similar constructions are performed for all anomaly categories, covering characteristic cues such as flames in Arson, standing water in Flood, crowd gathering in Riot, and high-speed motion in Traffic Accident. 
Besides hard normal samples, we also include ordinary normal videos generated from daily activities and reclassified normal samples identified during anomaly generation. All normal samples undergo the same component-aware verification procedure, ensuring that the distinction between normal and anomalous events is determined by event semantics rather than superficial visual patterns.

\subsection{Unified Multi-task Supervision}

All task annotations are derived from the same verified event representation, ensuring that different tasks share a consistent and reliable semantic foundation rather than being annotated independently. 
The benchmark organizes six tasks into three capability dimensions. Anomaly detection dimension includes conventional VAD-oriented detection (T1) and MLLM-oriented anomaly judgment (T3), both derived from the verified anomaly category. Anomaly retrieval and captioning dimension include video anomaly retrieval (T2), where the verified event components are organized into retrieval texts, and structured event captioning (T4). Fine-grained anomaly reasoning dimension consists of structured event reasoning (T5) and fine-grained question answering (T6), both constructed from the verified event semantics rather than generation prompts. 
To further evaluate semantic discrimination near the normal–anomaly boundary, T6 additionally contains boundary-sensitive multiple-choice questions (MCQs) covering both anomalous and hard normal samples. Together, the six tasks provide progressively richer supervision from anomaly detection to fine-grained reasoning while maintaining semantic consistency through the shared verified event representation.

\subsection{Dataset Statistics}

Table~\ref{tab:dataset_comparison} summarizes the overall statistics of AnomalyCraft-700K and compares it with existing video anomaly understanding benchmarks. AnomalyCraft-700K contains 29 anomaly categories and one normal category, comprising 40,100 videos and approximately 3.01 million frames. All videos are synthesized using Wan2.2 at a resolution of 1280×704 and 16 FPS, with durations ranging from 4 to 8 seconds. This duration covers both short-lived anomalies and more complex events involving multiple interactions and evolving outcomes. 
Beyond the video data itself, AnomalyCraft-700K provides rich multimodal supervision derived from the verified event representations. Across T3–T6, the benchmark contains 743,350 task-level language supervision instances, while T1 and T2 provide corresponding supervision for anomaly detection and video-text retrieval. All task annotations are generated from the same verified semantic representation, ensuring semantic consistency across different evaluation tasks. Detailed annotation statistics for each task are provided in the Appendix.

\section{Experiments}
\subsection{Experimental Setup}
We evaluate AnomalyCraft-700K on all six tasks. Samples are grouped by anomaly category, subject-object type, and generation source, and split as whole groups into training, validation, and test sets at a ratio of 70:5:25, so that samples from the same generation source do not appear across different splits. We conduct experiments on both task-specific deep models and MLLMs, with supervised fine-tuning (SFT) and zero-shot (ZS) evaluation on models with up to 4B parameters.
\subsection{Evaluation Metrics}
Higher values indicate better performance for all reported metrics. 
For anomaly detection (T1), we report threshold-independent AUROC and AP for binary detection, and Top-1 accuracy and macro-F1 for 30-way classification. For MLLM-oriented anomaly judgment (T3), we report binary accuracy, 30-way category accuracy, and category macro-F1. Bidirectional retrieval (T2) is measured by text-to-video and video-to-text Recall@1 and Recall@5. Structured event captioning (T4) uses BLEU-4, ROUGE-L, and CIDEr for the overall description, together with a Slot score, the mean BERTScore-F1 (B-F1) over the six event slots (context, subject, object, directed interaction, outcome, and background). For event-level reasoning (T5) and fine-grained QA (T6), outputs are scored by DeepSeek-V4-Pro, which is independent of both the annotation process and the evaluated models. Following the evaluation dimensions of VAU-R1, T5 is rated along classification (CLS), key-event matching (KM), fluency (FLU), informativeness (INF), and factual consistency (FAC), each from 0 to 10, while T6 open-ended answers are measured by BERTScore-F1 and judge correctness (Corr.). For multiple-choice QA in T6, we report accuracy and accuracy on the hard normal (HN) subset.

\subsection{Results on Anomaly Detection}
\label{sec:exp_recognition}

\begin{table}[t]
\centering

\small
\begin{tabular}{@{}lrrrr@{}}
\toprule
& \multicolumn{2}{c}{Binary(\%)} & \multicolumn{2}{c}{30-way(\%)} \\
\cmidrule(lr){2-3} \cmidrule(l){4-5}
Method & AUROC & AP & Top-1 & Macro-F1 \\
\midrule
RTFM~\shortcite{tian2021weakly}        & 92.43 & 97.12 & -- & -- \\
UR-DMU~\shortcite{zhou2023dual}      & 94.00 & 97.64 & -- & -- \\
BN-WVAD~\shortcite{zhou2024batchnorm}     & 94.52 & 97.96 & -- & -- \\
PEL4VAD~\shortcite{pu2024learning}     & 94.01 & 97.70 & -- & -- \\
JDC~\shortcite{kong2017joint}         & 92.63 & 96.81 & 58.46 & 50.84 \\
VadCLIP~\shortcite{wu2024vadclip}     & \textbf{96.82} & \textbf{98.72} & \textbf{73.94} & \textbf{69.81} \\
AnomalyCLIP~\shortcite{zhou2024anomalyclip} & 96.43 & 98.40 & 67.73 & 60.67 \\
DSANet~\shortcite{yin2026learning}      & 96.57 & 98.58 & 72.39 & 68.57 \\

\bottomrule
\end{tabular}
\caption{Video anomaly detection on AnomalyCraft-700K.}
\label{tab:t1_recognition}
\end{table}

\begin{table}[t]
\centering

\small
\begin{tabular}{@{}lrrr@{}}
\toprule
Model (set.)  & AUROC(\%) & Top-1(\%) & Macro-F1(\%) \\
\midrule
Qwen2.5-3B (ZS)  & 42.37 & 27.36 & 6.05 \\
Qwen2.5-3B (SFT) & 95.78 & 82.97 & 80.99 \\
Qwen3-4B (ZS)  & 86.33 & 28.48 & 11.91 \\
Qwen3-4B (SFT) & \textbf{96.81} & \textbf{87.13} & \textbf{85.07} \\
Qwen3-8B (ZS)  & 86.78 & 25.98 & 8.54 \\
InternVL3-8B (ZS) & 30.89 & 14.17 & 9.22 \\
\bottomrule
\end{tabular}
\caption{MLLM-oriented anomaly judgment on AnomalyCraft-700K.}
\label{tab:t3_judgment}
\end{table}

Table~\ref{tab:t1_recognition} reports the performance of conventional VAD methods on AnomalyCraft-700K. All baselines achieve strong binary detection results, with AUROC ranging from 92.43\% to 96.82\% and AP ranging from 96.81\% to 98.72\%. This indicates that the generated videos contain consistent and discriminative visual evidence of anomalous events. However, fine-grained recognition remains substantially more challenging: even the best-performing method reaches only 73.94\% Top-1 accuracy and 69.81\% macro-F1 under the 30-way setting. The large performance gap between binary detection and category recognition shows that identifying whether an event is anomalous is considerably easier than identifying its specific anomaly category. CLIP-based methods consistently outperform the conventional classifier JDC in 30-way recognition, further suggesting that language-aligned semantic representations are beneficial for distinguishing fine-grained anomaly categories beyond generic motion and appearance cues.

A similar pattern is observed for MLLM-oriented anomaly judgment in Table~\ref{tab:t3_judgment}. Under zero-shot evaluation, Qwen3 models already achieve over 86\% binary accuracy, yet all models remain below 30\% in category accuracy, revealing a clear discrepancy between coarse anomaly awareness and fine-grained semantic discrimination. Supervised fine-tuning on AnomalyCraft-700K substantially narrows this gap: category accuracy increases from 27.36\% to 82.97\% for Qwen2.5-3B and from 28.48\% to 87.13\% for Qwen3-4B, accompanied by similarly large macro-F1 improvements. Moreover, the zero-shot Qwen3-8B does not outperform Qwen3-4B in category recognition, indicating that model scaling alone is insufficient to acquire domain-specific anomaly semantics. Together, T1 and T3 demonstrate that AnomalyCraft-700K provides reliable anomaly evidence while offering the fine-grained supervision required to transform coarse anomaly detection into category-level semantic understanding.

\subsection{Results on Anomaly Retrieval and Captioning}
\label{sec:exp_association}

\begin{table}[t]
\centering

\small
\begin{tabular}{@{}lrrrr@{}}
\toprule
& \multicolumn{2}{c}{T$\rightarrow$V(\%)} &
  \multicolumn{2}{c}{V$\rightarrow$T(\%)} \\
\cmidrule(lr){2-3} \cmidrule(l){4-5}
Method & R@1 & R@5 & R@1 & R@5 \\
\midrule
CLIP4Clip~\shortcite{luo2022clip4clip}& 85.36 & 97.52 & 87.48 & 97.89 \\
X-CLIP-B/32~\shortcite{ma2022x}       & 85.42 & 97.53 & 86.29 & 97.73 \\
X-CLIP-B/16~\shortcite{ma2022x}       & \textbf{87.99} & \textbf{98.37} &
                     \textbf{87.82} & \textbf{98.17} \\
TS2-Net~\shortcite{liu2022ts2}           & 84.31 & 97.54 & 87.28 & 97.92 \\
UCoFiA~\shortcite{wang2023unified}            & 86.89 & 97.87 & 87.32 & 98.13 \\
EagleNet~\shortcite{chen2026eaglenetenergyawarefinegrainedrelationship}          & 82.10 & 96.31 & 82.84 & 96.71 \\
\bottomrule
\end{tabular}
\caption{Video anomaly retrieval on AnomalyCraft-700K.}
\label{tab:t2_retrieval}
\end{table}

\begin{table}[t]
\centering

\small
\begin{tabular}{@{}lrrrr@{}}
\toprule
Model (set.) & B-4 & R-L & CIDEr & Slot \\
\midrule
Qwen2.5-3B (ZS)   & 0.0479 & 0.2247 & 0.7825 & 0.7215 \\
Qwen2.5-3B (SFT)  & 0.1190 & 0.3219 & 1.4024 & 0.7853 \\
Qwen3-4B (ZS)     & 0.0515 & 0.2238 & 0.8892 & 0.7160 \\
Qwen3-4B (SFT)    & \textbf{0.1384} & \textbf{0.3459} &
                     \textbf{1.6406} & \textbf{0.7912} \\
Qwen3-8B (ZS)     & 0.0519 & 0.2217 & 0.9056 & 0.7153 \\
InternVL3-8B (ZS) & 0.0485 & 0.2380 & 0.9246 & 0.7332 \\
\bottomrule
\end{tabular}
\caption{Structured event captioning on AnomalyCraft-700K.}
\label{tab:t4_caption}
\end{table}

\begin{table*}[t]
\centering

\small
\begin{tabular*}{\textwidth}{@{\extracolsep{\fill}}llrrrrrr@{}}
\toprule
Model & Setup & CLS & KM & FLU & INF & FAC & Total \\
\midrule
Qwen2.5-3B  & ZS  & 2.03 & 2.42 & 6.50 & 2.47 & 2.65 & 16.06 \\
Qwen2.5-3B  & SFT & 8.34 & 6.82 & 8.60 & 7.03 & 6.78 & 37.57 \\
Qwen3-4B    & ZS  & 5.96 & 5.24 & 8.23 & 5.49 & 5.01 & 29.93 \\
Qwen3-4B    & SFT & \textbf{8.77} & \textbf{7.80} & \textbf{8.93} &
                       \textbf{7.95} & \textbf{7.77} & \textbf{41.22} \\
Qwen3-8B    & ZS  & 6.53 & 5.60 & 8.35 & 5.87 & 5.25 & 31.59 \\
InternVL3-8B & ZS & 3.17 & 4.40 & 7.30 & 3.99 & 4.85 & 23.72 \\
\bottomrule
\end{tabular*}
\caption{Structured event reasoning on AnomalyCraft-700K.}
\label{tab:t5_analysis}
\end{table*}

\begin{table}[t]
\centering

\small
\begin{tabular}{@{}lrrrr@{}}
\toprule
& \multicolumn{2}{c}{QA} & \multicolumn{2}{c}{MCQ} \\
\cmidrule(lr){2-3} \cmidrule(l){4-5}
Model/set. & B-F1 & Corr. & Acc. & HN Acc. \\
\midrule
Q2.5-3B/ZS  & 0.7048 & 0.2064 & 0.8480 & 0.7362 \\
Q2.5-3B/SFT & 0.7883 & 0.5816 & 0.9872 & 0.9858 \\
Q3-4B/ZS    & 0.6983 & 0.3559 & 0.8060 & 0.7455 \\
Q3-4B/SFT   & \textbf{0.7952} & \textbf{0.7522} &
               \textbf{0.9904} & \textbf{0.9881} \\
\bottomrule
\end{tabular}
\caption{Fine-grained question answering on AnomalyCraft-700K.}
\label{tab:t6_qa}
\end{table}

Table~\ref{tab:t2_retrieval} reports video anomaly retrieval performance on AnomalyCraft-700K. Across diverse retrieval architectures, T$\rightarrow$V R@1 ranges from 82.10\% to 87.99\%, while V$\rightarrow$T R@1 ranges from 82.84\% to 87.82\%. The consistently strong bidirectional performance indicates reliable global correspondence between the generated videos and their verified textual annotations. X-CLIP-B/16 achieves the best results in both directions, outperforming X-CLIP-B/32 by 2.57\% and 1.53\% on T$\rightarrow$V and V$\rightarrow$T R@1, respectively. This suggests that finer visual representations better capture anomaly-specific entities, interactions, and contextual cues. Overall, these results primarily validate the reliability of the verified video–text alignment, as T2 is intended to assess supervision quality rather than retrieval difficulty alone.

Table~\ref{tab:t4_caption} further evaluates whether this global correspondence transfers to structured event captioning. Supervised fine-tuning consistently improves all metrics for both Qwen model families. In particular, Qwen3-4B improves CIDEr from 0.8892 to 1.6406 and the Slot score from 0.7160 to 0.7912. By contrast, scaling the zero-shot model from 4B to 8B provides little benefit and slightly reduces the Slot score, again indicating that task-specific supervision is more effective than model scaling alone. The supplementary slot-level results show the largest gains for object, subject, and background, demonstrating improved grounding of individual event components. Together, T2 verifies reliable global video–text association, while T4 shows that AnomalyCraft-700K supervision supports fine-grained grounding and description of event components rather than merely learning a template-like output.

\subsection{Results on Fine-grained Anomaly Reasoning}
\label{sec:exp_reasoning}

Table~\ref{tab:t5_analysis} reports the results on structured event reasoning (T5). Across all five evaluation dimensions, supervised fine-tuning consistently yields substantial improvements over zero-shot inference. The total score increases from 29.93 to 41.22 for Qwen3-4B and from 16.06 to 37.57 for Qwen2.5-3B. Notably, fluency is already the strongest dimension under zero-shot settings, with scores above 8 for most models, and exhibits only modest improvement after fine-tuning. In contrast, classification, key-event matching, informativeness, and factual consistency all improve markedly, indicating that AnomalyCraft-700K primarily enhances semantic understanding and structured reasoning rather than surface-level language generation. Moreover, the zero-shot Qwen3-8B achieves only 31.59, still substantially below the fine-tuned 4B model, suggesting that high-quality task supervision contributes more to fine-grained anomaly reasoning than simply increasing model scale.

The same trend is observed on fine-grained question answering (T6), as shown in Table~\ref{tab:t6_qa}. QA correctness increases substantially after fine-tuning, increasing from 0.3559 to 0.7522 for Qwen3 and from 0.2064 to 0.5816 for Qwen2.5, while the corresponding B-F1 gains are much smaller. This discrepancy indicates that lexical similarity alone cannot adequately measure semantic correctness, further highlighting the necessity of reasoning-oriented evaluation. For multiple-choice QA, accuracy reaches 0.9904 and 0.9872 after fine-tuning, demonstrating that the proposed annotations effectively support semantic discrimination. More importantly, the largest improvements are consistently observed on the hard normal subset, where accuracy rises from 0.7455 to 0.9881 for Qwen3 and from 0.7362 to 0.9858 for Qwen2.5. This confirms that the proposed hard normal samples constitute an effective test of semantic boundary understanding, particularly in the zero-shot setting, requiring models to distinguish visually similar events according to their underlying semantics rather than superficial visual cues.

\section{Conclusion}
We present AnomalyCraft-700K, a component-controllable synthetic benchmark for fine-grained video anomaly understanding. To address the limited anomaly priors of general-purpose video generators, we develop a generation scheme suited to fine-grained anomalies through component-level anomaly representation and a progressively refined three-stage pipeline, thereby crafting anomaly videos with fine-grained semantics and constructing per-category hard normal samples that are visually close to anomalies yet semantically normal; on this basis, we verify each component to ensure the component-level accuracy and cross-modal consistency of the annotations. Experiments across six tasks show that AnomalyCraft-700K provides effective semantic supervision for models ranging from anomaly detection to fine-grained anomaly understanding. Future work will explore AnomalyCraft-700K as a general-purpose anomaly benchmark, evaluating whether pretraining or joint training on it improves cross-dataset generalization and complements existing real-world anomaly benchmarks across broader video anomaly understanding tasks.

\appendix

\bibliography{aaai2027}

\end{document}